\documentclass[10pt,twocolumn,letterpaper]{article}

\usepackage[pagenumbers]{cvpr} 

\usepackage{graphicx}
\usepackage{amsmath}
\usepackage{amssymb}
\usepackage{booktabs}
\usepackage[inline]{enumitem}
\usepackage{multirow}
\usepackage{threeparttable}
\usepackage{ntheorem}
\usepackage{footnote}
\makesavenoteenv{tabular}
\makesavenoteenv{table}
\usepackage{enumitem}
\usepackage{url}
\theoremseparator{:}

\usepackage[pagebackref,breaklinks,colorlinks]{hyperref}

\usepackage[capitalize]{cleveref}
\crefname{section}{Sec.}{Secs.}
\Crefname{section}{Section}{Sections}
\Crefname{table}{Table}{Tables}
\crefname{table}{Tab.}{Tabs.}

\def\cvprPaperID{*****} 
\def\confName{CVPR}
\def\confYear{2022}

\begin{document}

\title{Factors that affect Camera based Self-Monitoring of Vitals in the Wild}

\author{Nikhil S. Narayan \\
\and
Shashanka B. R. \thanks{The authors were at MFine when the research was carried out. Prior consent has been obtained from the authors to publish this work}\\
\and
Rohit Damodaran$^*$\\
\and
Dr. Chandrashekhar Jayaram$^*$\\
\and
Dr. M. A. Kareem\\
\and
Dr. Mamta P.$^*$\\
\and
Dr. Saravanan K. R.$^*$\\
\and
Dr. Monu Krishnan$^*$\\
\and
Dr. Raja Indana\\\\
MFine\\
}
\maketitle

\begin{abstract}
The reliability of the results of self monitoring of the vitals in the wild using medical devices or wearables or camera based smart phone solutions is subject to variabilities such as position of placement, hardware of the device and environmental factors. In this first of its kind study, we demonstrate that this variability in self monitoring of Blood Pressure (BP), Blood oxygen saturation level (SpO2) and Heart rate (HR) is statistically significant ($p<0.05$) on 203 healthy subjects by quantifying positional and hardware variability. We also establish the existance of this variability in camera based solutions for self-monitoring of vitals in smart phones and thus prove that the use of camera based smart phone solutions is similar to the use of medical devices or wearables for self-monitoring in the wild.  

\end{abstract}

\section{Introduction}
\label{sec:intro}

Monitoring of vitals is important for identifying and managing of diseases. Glasziou et. al. \cite{glasziou2005monitoring} identify five phases of monitoring a disease from a chronic condition perspective along with the interval for monitoring. These include continuous monitoring of vitals to (a) detect abnormality; (b) confirm abnormality; (c) establish plan of action to treat the abnormality; (d) make adjustments to the treatment; and (e) to confirm the success of treatment. These phases can very well be extended to acute conditions without loss of generality except for the fact that the monitoring intervals may be much shorter than that for chronic conditions and may totally be based on self-monitoring. Self-monitoring refers to monitoring of conditions by patients/users by using medical devices that are available off-the-shelf at a place of the patient's convenience without the intervention of a qualified medical professional to monitor the condition. Self monitoring in the wild refers to monitoring the vitals in uncontrolled environments such as outdoors or house where the conditions with respect to lightihg, physical activity etc., dynamically change.


At certain points in time, when access to care is made difficult either due to the unavailability of care providers or due to environmental factors such as the most recent COVID-19 pandemic where patients were forced to monitor their conditions at home before approaching a healthcare provider, self-monitoring plays a crucial role in saving human lives. Given the importance of self-monitoring  of diseases in recent times, there has been a flurry of activity in research circles to come up with novel technologies that enable self-monitoring of vitals such as Blood Pressure (BP), Blood oxygen saturation level (SpO2), Heart rate (HR), Blood Glucose etc. While some of these have made it to commercial devices such as wearables and mobile phones, a vast majority of the technology is still restricted to academia and research circles.

\subsection{Prior art}
\label{ssec:PA}
Of the technologies that are being researched for self-monitoring of vitals, computer vision has received considerable amount of attention the the last decade. One of the early attempts in this direction was by Jonathan et. al., \cite{jonathan2010investigating} where the authors used a Nokia device to record videos at 15 FPS of a user before and after performing a physical activity and the change in HR calculated by employing using Photoplethysmography(PPG) Imaging. Thereafter there have been quite a few attempts to demonstrate the capability of using PPG signals from videos/image stream to estimate HR in an individual \cite{matsumura2013iphysiometer, gregoski2011photoplethysmograph, kwon2012validation, gregoski2012development, chatterjee2018ppg, siddiqui2016pulse}. 

The PPG imaging methodology employed for HR is also a popular method to estimate SpO2 by both signal processing/mathematical models \cite{scully2011physiological,lamonaca2015blood, ding2018measuring} and machine learning \cite{krizhevsky2012imagenet, carni2017blood, kanva2014determination, demlabim2022smartphox, carni2016setting}. A majority of smart phone PPG solutions for SpO2 consider the video of a finger tip placed against the back camera of the phone, hereafter referred to as Finger tip (FT) PPG, as the input to extract the PPG signals and estimate the vital. A non-contact way to monitor SpO2 is explored in \cite{guazzi2015non, wei2021analysis}where a video stream of user's face, hereafter referred to as Face (FC) PPG, is used as input for PPG signal extraction and SpO2 Estimation.

There has been growing interest in recent times to estimate BP using PPG signals extracted either from FT PPG or FC PPG. While, the exact relationship between PPG the signal and Blood Pressure has not been clinically established, fitting an ML model on top of either the raw signal, or some features extracted from the PPG signal seems to work and is the general direction of work so far.  Neural Networks \cite{slapnivcar2019blood,lamonaca2013application, 7590775, baek2020blood}, Ensemble methods\cite{gaurav2016cuff} and LSTM's \cite{radha2019estimating} are some popular machine learning algorithms that have been employed to estimate BP using either the raw PPG signal or features extracted from the raw PPG signal as inputs to the algorithms. Efforts towards developing contact-less solutions include using a camera to capture facial videos of the user and use deep Learning models on the extracted rPPG(remote PPG) signals to estimate BP \cite{schrumpf2021assessment}. Transdermal Optical Imaging technology is another way to capture facial blood flow changes and estimate BP in a contact-less manner \cite{luo2019smartphone}. 


\subsection{Motivation}
\label{ssec:Mot}
The algorithms discussed in Section \ref{ssec:PA} either use publicly available datasets or data available from a limited set of devices (usually one) to develop camera based vitals monitoring solutions. PhysioNet \cite{goldberger2000physiobank} is a popular database commonly used to develop algorithms based on Finger tip PPG. Algorithms based on Face PPG use either \cite{estepp2014recovering} or \cite{zhang2016multimodal} or both to train the models. It is interesting to note that these datasets are constructed under strictly controlled environment such as fixed lighting settings, fixed background, fixed distance from camera etc., and do not account for interdevice variability. Thus, these datasets do not account for variabilities that arise when the devices are used in the wild or when multiple devices of different kinds are used for self-monitoring. Some examples of variabilities that may arise while acquiring data for self monitoring in the wild are:
\begin{itemize}
    \item \emph{Positional Variance}: (a) Wrong positioning of the BP cuff of a digital monitor during data acquisition for BP algorithms. Previous studies have established that there is a significant Inter Arm BP difference when a digital monitor is used for BP measurement \cite{lane2002inter,orme1999normal}. Several clinical studies in the past have also established the existence of variabilities in the technique used to measure BP by experts and the negative impact that it has while monitoring in a clinical setting \cite{villegas1995evaluation,schulze2002effect,ray2012blood}. When such variabilities exist in measurements taken by experts themselves, it is not uncommon to expect the variability to exist while measuring BP in a self monitoring setting. (b) Similar positional variances exist while measuring SpO2 and HR using pulseoximeters.
    \item \emph{Hardware Variance}: Variabilities in the hardware used to acquire Image / Video signals for PPG based vitals analysis. It is a well known fact that manufacturers of different brands of smart phones employ sensors of different Original Equipment Manufacturers (OEM's) to have a competitive edge. This will directly result in the variability of the quality of PPG signal obtained from different sensors.
    \item \emph{Environmental Variance}: the mobility aspect of smart phones by default will introduce variances related to lighting, motion etc.
\end{itemize}
\begin{table}[!t] \footnotesize
\centering
\renewcommand{\arraystretch}{1.3}
\begin{threeparttable}
\caption{Mobile applications and wearables that support vitals monitoring}
\begin{tabular}[!t]{c c c c c c }\hline
\multirow{2}{*}{\bf Technology} & \bf \multirow{2}{*}{\bf Name} & \multicolumn{3}{c}{\bf Vitals support} & \bf \multirow{2}{*}{\bf Solution}\\
& & BP & SpO2 & HR & \\\hline
Wearable & Apple Watch &  & Y & Y & IoT\\
 & GOQii & Y & Y & Y & IoT\\
 & boAT Xtend &  & Y & Y & IoT\\
 & One Plus &  & Y & Y & IoT\\
 & Fitbit &  & Y & Y & IoT\\
 & Oura ring &  &  & Y & IoT\\
 & Omron & Y &  &  & IoT\\\cline{2-6}
 Mobile App& Careplix &  & Y & Y & Camera (FT)\\
 & MFine  & Y & Y & Y & Camera (FT)\\
 & ICICI  & Y & Y & Y & Camera (FC)\\
 & Anura & Y &  & Y & Camera (FC)\\\hline
\end{tabular}
\label{tab: App_Comp}
\end{threeparttable}
\end{table}

In this paper we address the following questions quantitatively/statistically by undertaking a systematic clinical study: \begin{enumerate*}[label=(\alph*)]
    \item What variabilities exist in self monitoring of the following vitals: BP, SpO2 and HR? 
    \item How do these variabilities compare to the measurements obtained by an expert who is a qualified medical professional?
    \item If the variabilities between self-monitoring and the measurements by an expert are similar, then how do these variabilities affect the training of computer vision based solutions? for vitals monitoring?
    \item What do we need to do to minimise the effect of this variability in AI based solutions?
\end{enumerate*}

The focus of this paper is on the variabilities associated with Position and Hardware while Environmental Variance will be picked up as an extension to this study to do justice to the number of environmental factors that may influence the outcome of self-monitoring in the wild. It should also be noted that the primary goal of this paper is to validate commercially available solutions at this point in time as we wish to evaluate that solution which is easily accessible to a user in the current situation given the pandemic and the global burden on healthcare infrastructure.

\subsection{Contributions}
\label{ssec:Contrib}

To the best of our knowledge, there is no prior work that has formally quantified the variabilities that exist in self-monitoring of vitals. Since a majority of solutions in literature are targeted towards self-monitoring of vitals, it is crucial to identify this variability and determine it's significance in order to establish error bounds for measurements of vitals on smartphones and wearables in the future. In this first-of-a-kind study, we establish statistically that there is a significant variability in the Vitals when measured by self using medical devices that are available off the shelf.
Additionally, we also establish the existance of this variability in camera based solutions for self-monitoring of vitals and thus prove that the use of camera based smart phone solutions is similar to the use of medical devices or wearables for self-monitoring. Finally, discuss some methods that can potentially be used to reduce this variability in camera based solutions.

\begin{figure}
    \centering
    \includegraphics[width=2.7in]{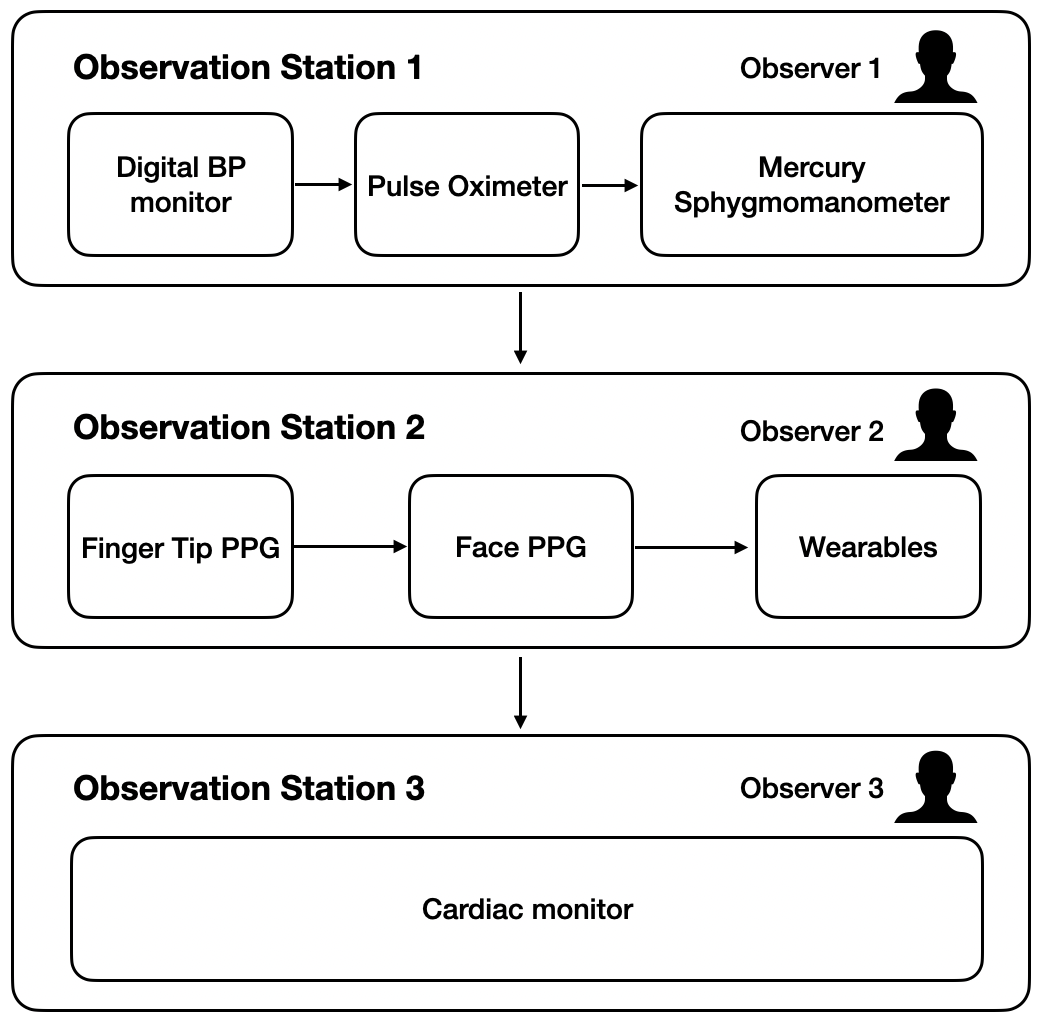}
    \caption{Study setup}
    \label{fig:Obs_Stn}
\end{figure}

\section{Materials and Method}

\subsection{Data}
\label{ssec:Data}

The study was conducted on 203 healthy subjects comprising of 112 (55\%) men and 91 (45\%) women in the age range [20,55] years with an average age of $28.77 \pm 5.877$ years. The sample size was estimated for a 5\% error at 95\% confidence interval based on the computations in \cite{fonseca2012simultaneous,lane2002inter,mendelson2004simultaneous}. Prior written consent was obtained from the subjects who volunteered for the study as per the IRB guidelines and approvals obtained for the study.

In order to monitor the vitals on medical devices, mobile phones and wearables, we will require at least: 
\begin{enumerate}[label*=\arabic*.]
    \item one pair of mobile phones/smartphones with one of the phones pre-installed with a mobile application that measures vitals by extracting PPG signals from finger tip image stream/video and the other from face videos from front camera of the phone. These mobile applications are assumed to employ any one of the methods described in Section \ref{ssec:PA}. The exact details on the nature of the algorithm are unavailable at this point in time as the developers of the applications have not disclosed it publicly as either patents or publications.
        \subitem  The phones of the pair should be of the same brand to eliminate the variabilities that may arise due to differences in the hardware and software used to acquire the signals (Image/Video). Since the study also involves observing the variabilities associated with changes in hardware and software of the smartphones, we employ 4 pairs of phones in this study. The criteria for selection of the phone brand/model is based on price of the phone and global availability of the phone. Lower the price, higher is the reach to the people who would need access to affordable healthcare. The price range under consideration for this study is US\$100-US\$250. Accordingly, the following 4 models of phones are used: \begin{enumerate*}[label*=(\alph*)]
            \item Xiaomi Note 9 Pro (Xi N9);
            \item Xiaomi Note 8 Pro (Xi N8);
            \item Oppo A15 (Oppo); and
            \item Samsung M31 (SM31);
        \end{enumerate*}
 
        \subitem The mobile applications should be capable of measuring all the vitals that are considered for this study. Table \ref{tab: App_Comp} shows different commercially available mobile applications and wearables along with the vitals supported by each. We select one mobile application each for Finger tip PPG based monitoring and Face PPG based monitoring from the Mobile App category in Table \ref{tab: App_Comp}. Accordingly, the following mobile applications are considered for this study: MFine for Finger tip PPG based monitoring and ICICI mobile application for Face PPG based monitoring.
    
    \item one wearable. As with the mobile phones, the wearable selected for this study should cover all the vitals considered for the study. Froom Table \ref{tab: App_Comp} it can be seen that the GOQii smartwatch supports all vitals and is thus used in the study. Since we also want to study the variabilities between wearables, we have also included Apple Watch Series 7 in our study even though BP monitoring capability is absent in it;
    \item a digital BP monitor. Omron HEM 7121J fully automatic Digital Blood Pressure monitor is used in this study; 
    \item a mercury based sphygmomanometer. Diamond Mercurial Blood Pressure apparatus was used for this study;
    \item a pulseoximeter. BPL Smary Oxy pulseoximeter was used for this study to measure SpO2 and Heart Rate; and
    \item a cardiac monitor that will be used as gold standard to eliminate the interobserver variability that may arise with manual measurements with a sphygmomanometer. Yonker YK 8000C Multi-parameter patient monitor was used in this study.
\end{enumerate}


\begin{figure}
    \centering
    \includegraphics[width=3.0in]{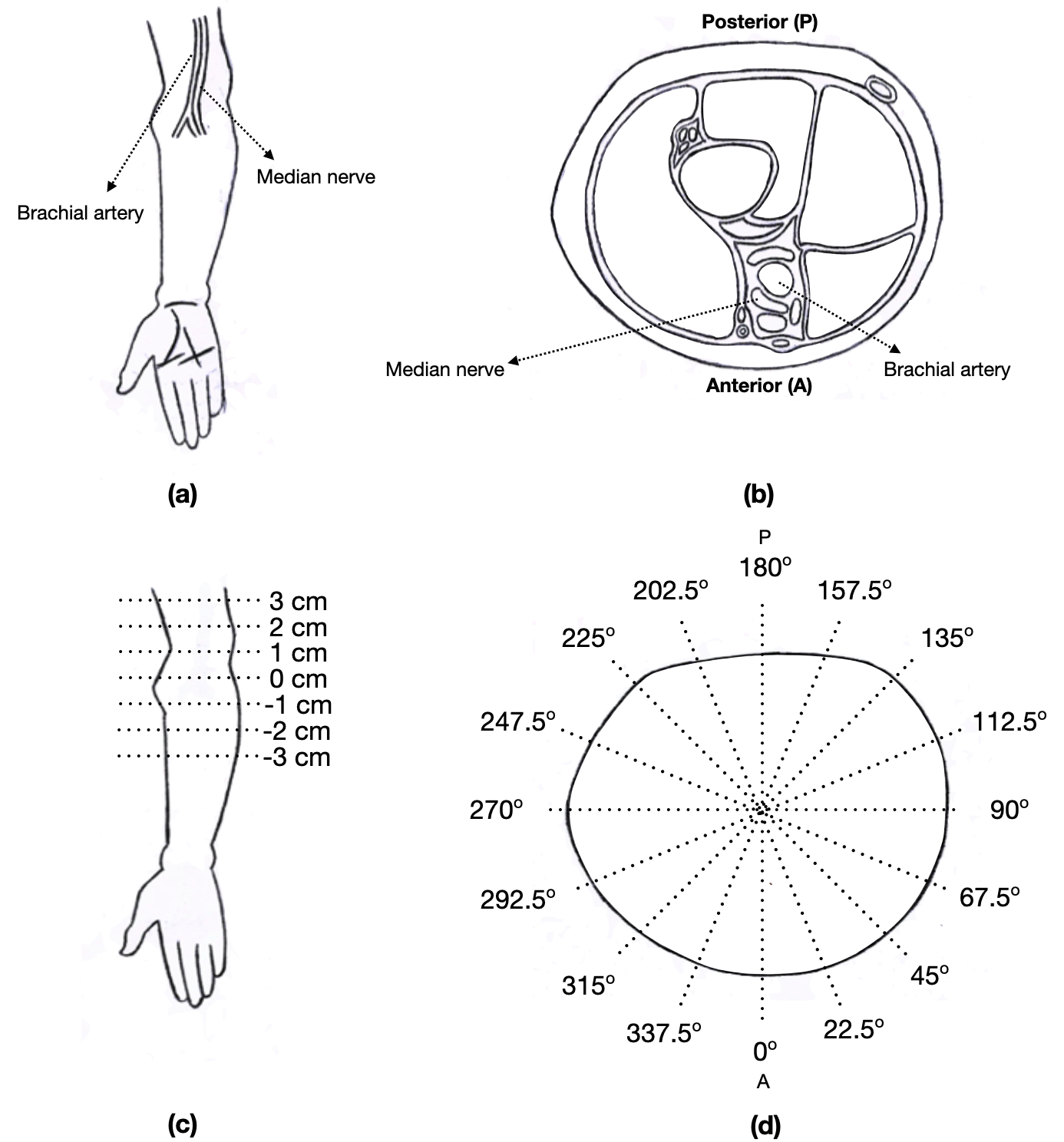}
    \caption{Quantifying the (a,c) transverse variability; and (b,d) angular variability in the placement of the sensor of the BP cuff for self-monitoring}
    \label{fig:BP_Anatomy}
\end{figure}

\begin{figure}
    \centering
    \includegraphics[width=3.0in]{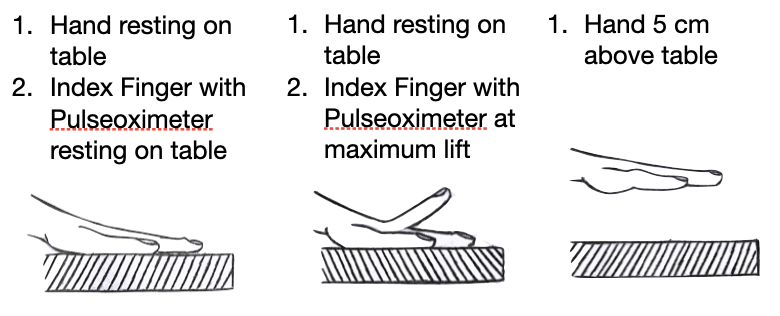}
    \caption{Variations in measuring SpO2 and HR by placing a pulseoximeter on the index finger of the hand (a) resting on the table; (b) resting on the table with index figer at maximum angle; and (c) in the air. }
    \label{fig:SpO2_Anatomy}
\end{figure}

\begin{figure}
    \centering
    \includegraphics[width=3.0in]{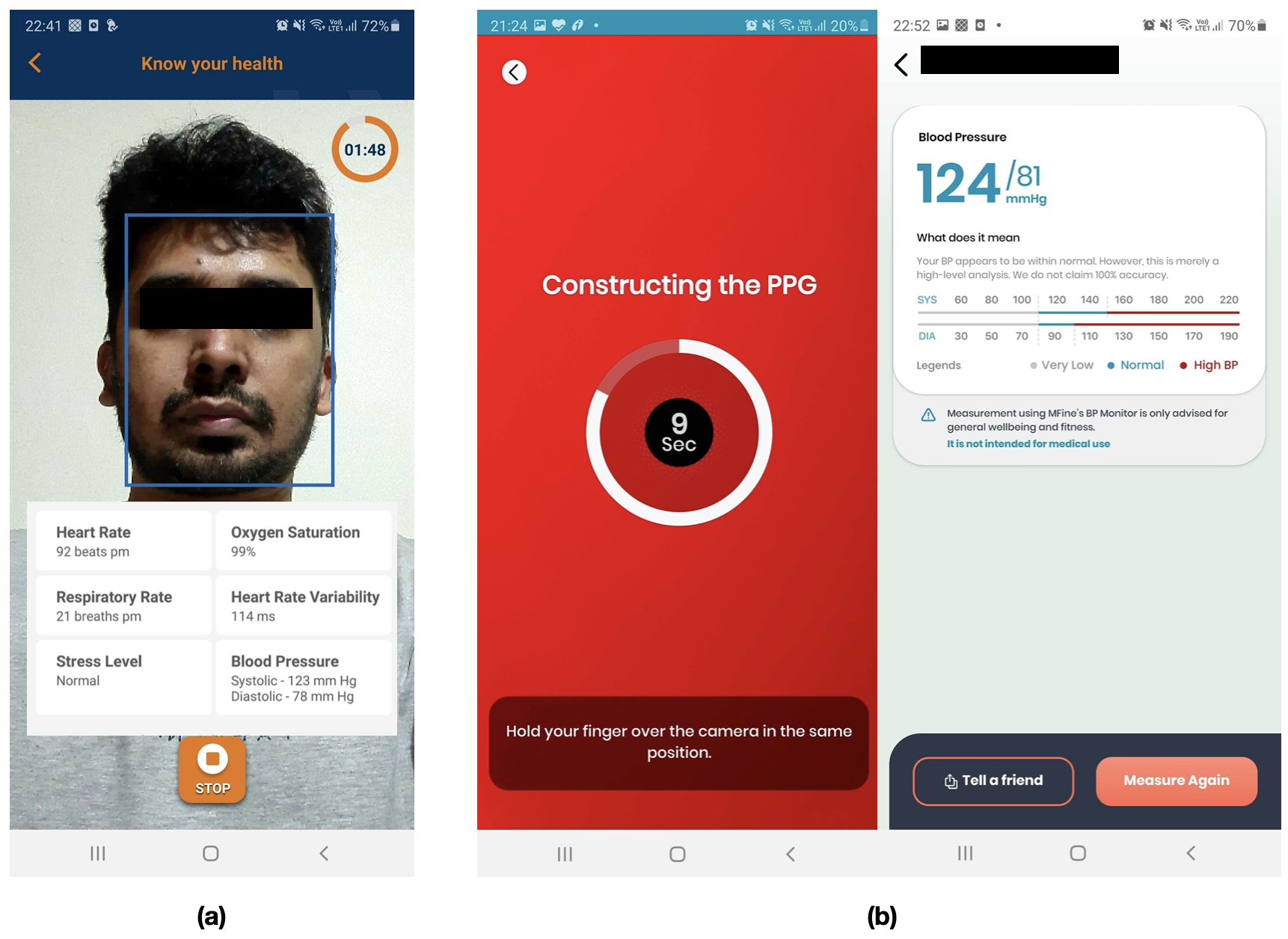}
    \caption{Commercially available vitals monitoring solutions that use (a) face videos (selfie); and (b) Finger tip videos to measure vitals such as BP, SpO2 and HR.}
    \label{fig:PPG_Illustration}
\end{figure}




\subsection{Methodology}
The study setup comprises of three observation stations in a well lit room manned by three independent observers who are qualified medical practitioners for : \begin{enumerate*}[label=(\alph*)]
    \item self monitoring;
    \item monitoring with camera based mobile application and wearables; and
    \item monitoring with a cardiac monitor
\end{enumerate*}. Figure \ref{fig:Obs_Stn} illustrates the setup used for this study.

The first station is where the variability in self monitoring is studied. The self monitoring exercise starts with the subjects being asked to measure BP using a digital BP monitor that is placed in front of him/her without any instructions given on how to operate the BP monitor (which also includes the placement of the cuff on the arm). The observer of the station then notes down the position of the sensor in the cuff on the arm with respect to: (a) the displacement along the arm which is quantified as per the grading in Fig \ref{fig:BP_Anatomy} (c); and (b)  the approximate angle it makes with the centreline (0$^\text{o}$) of Fig \ref{fig:BP_Anatomy} (d). Once the BP measurement and the corresponding variations are recorded by the observer, the subjects are then asked to measure the SpO2 and HR by placing the pulseoximeter on the index finger and starting the measurement. The observer notes down the SpO2 and HR readings after 15s of the start of the measurement at each of the positions indicated in Fig \ref{fig:SpO2_Anatomy}. The last step in station 1 involves the observer measuring the Blood pressure of the subject using a mercury based sphygmomanometer.

\begin{figure}
    \centering
    \includegraphics[width=3in]{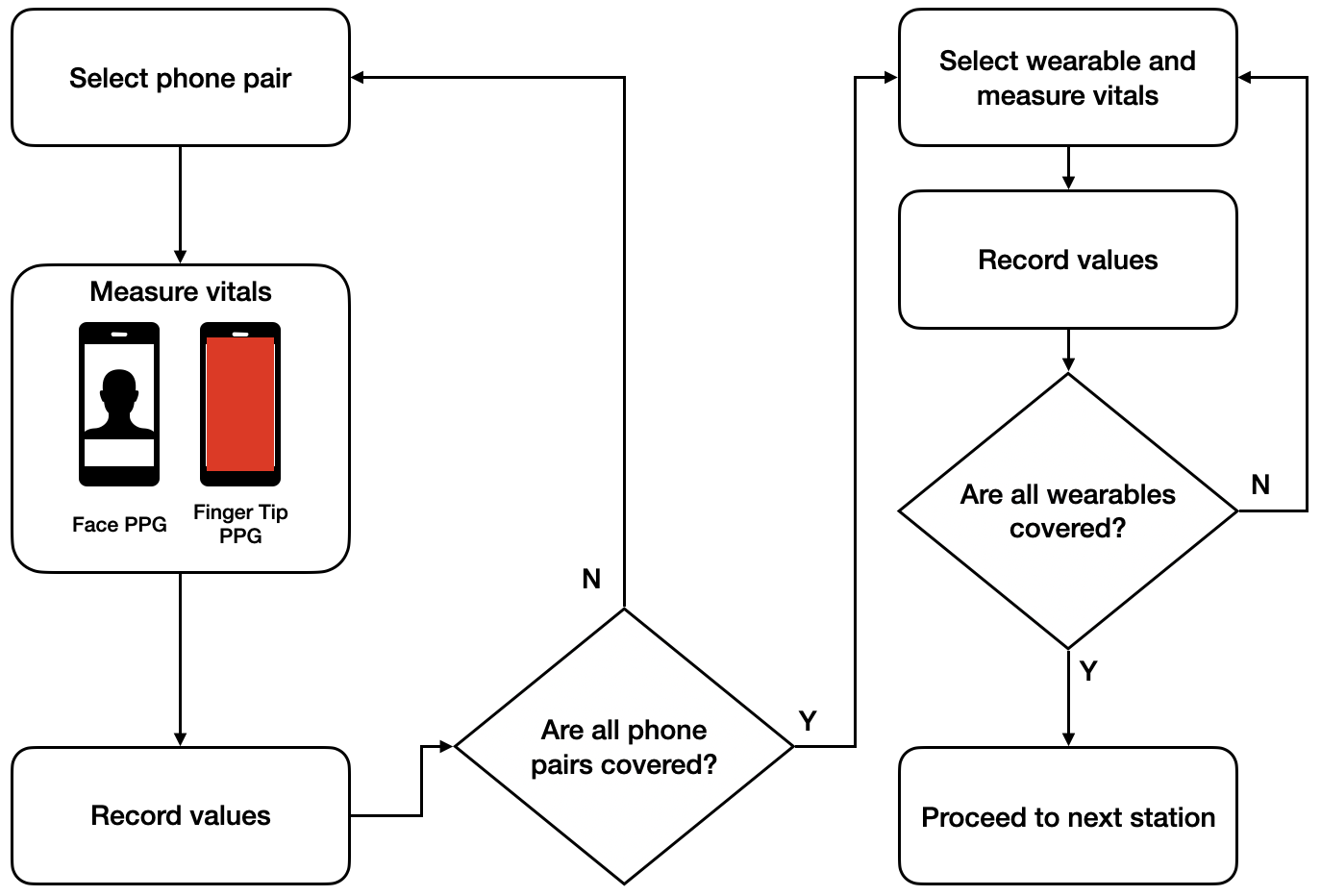}
    \caption{Vitals monitoring using mobile phones and wearables in the 2nd station}
    \label{fig:Obs_Stn_2}
\end{figure}

The subject is then asked to proceed to the second station where the vitals are monitored using mobile based applications and wearables. As described in Section \ref{ssec:Data}, we use 4 pairs. of mobile phones of and two wearables per subject to monitor the vitals. One phone of each pair is used for vitals monitoring using Face PPG and the other is used for vitals monitoring using Finger Tip PPG. Phones in each pair belong to the same brand and measurements on both phones are taken simultaneously. The phone used for Face PPG is placed on a mobile phone holder with the angle adjusted so that the front camera points to the face of the subject (with the full face in view). Figure \ref{fig:PPG_Illustration} illustrates the face based and finger tip based methods to measure Vitals on mobile phones. Figure \ref{fig:Obs_Stn_2} shows the procedure followed to measure the vitals in the second observation station. Subjects who have completed measuring their vitals in the second station are asked to proceed to the final observation station where their vitals are monitored by a cardiac monitor. The measurements of the final station are used as the gold standard for evaluating the measurements obtained at each observation station.

\section{Experimental results}
In this section we demonstrate the variabilities that exist in self monitoring using medical devices, mobile phones and wearables, respectively. We use oneway ANOVA and one tailed Student's t-test to establish the statistical significance of the variabilities using p-values. Where variabilities are concerned, we use the mean and variance of differences between the measurements of the device under consideration and the gold standard which is the cardiac monitor.
\begin{figure*}
    \centering
    \includegraphics[width=6.5in]{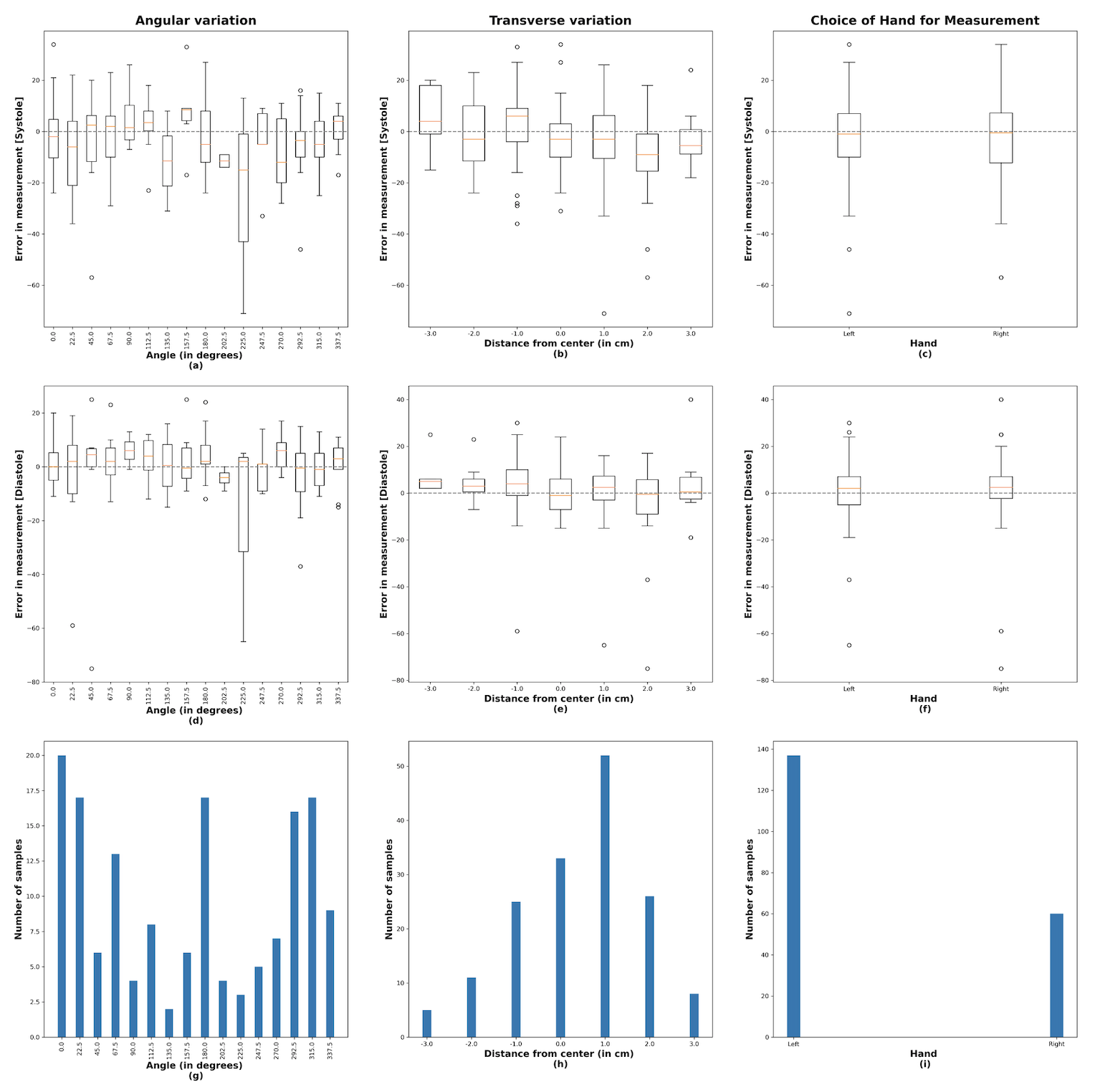}
    \caption{Variability in self monitoring of Blood Pressure (BP) using digital BP monitors. (a,d,g). Variations in the angular placement of the sensor (in the cuff) on the arm near the synovial joint; (b,e,h). Variations in the placement of the sensor (in the cuff) along the arm near the synovial joint; (c,f,i). Variations in the selection of the arm [Left/Right] to place the sensor for BP measurement.}
    \label{fig:BP_Variations}
\end{figure*}

\subsection{Self monitoring of vitals using medical devices}
\paragraph{BP} The plots in Figure \ref{fig:BP_Variations} show the variability in both Systolic and Diastolic BP measurements when the sensor in the cuff of the digital BP monitor is placed at different angles around the arm and at different positions along the arm (Left/Right). It is interesting to note that the Systole BP measurements obtained on the Left hand showed a statistically significant difference ($p<0.05$) with the readings obtained from the cardiac monitor. Difference in measurements between cardiac monitor and digital BP monitor for Systolic BP measurement was statistically significant ($p<0.05$) while Diastolic measurements were within statistically acceptable limits ($p>0.05$). Both Systolic and Diastolic BP readings did not show statistically significant results ($p>0.05$) when compared to the measurement by an expert using mercury based Sphygmomanometer.

\begin{figure}
    \centering
    \includegraphics[width=2.5in]{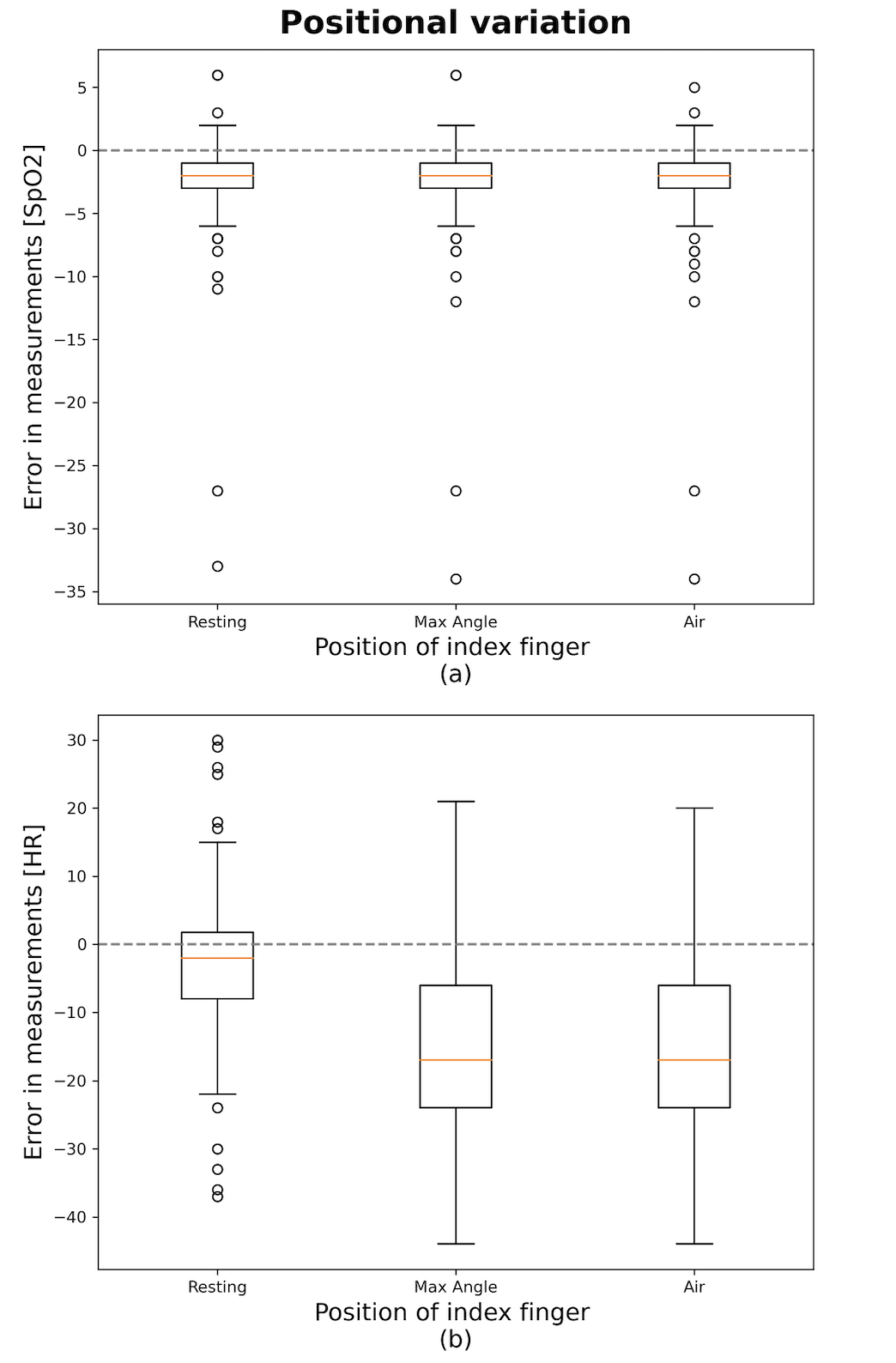}
    \caption{Positional variability in measuring SpO2 and HR using pulseoximeter}
    \label{fig:SpO2_Variations}
\end{figure}

\paragraph{SpO2 and HR} The plots in Figure \ref{fig:SpO2_Variations} shows the variability in the measurements of both SpO2 and HR when the pulseoximeter is clamped on to the index finger of the hand according to the variations shown in Fig. \ref{fig:SpO2_Anatomy}.

Both SpO2 and Heart Rate showed statistically significant difference ($p<0.05$) between pulseoximeter and cardiac monitor readings at all positions. A one way ANOVA performed on the pulseoximeter readings at different positions of hand and index finger showed no statistically significant difference ($p>0.05$) between the readings for SpO2. Heart Rate on the other hand showed a statistically significant difference ($p<0.05$) between the readings at different positions.

\subsection{Self monitoring of vitals using mobile devices}

\begin{figure*}
    \centering
    \includegraphics[width=6in]{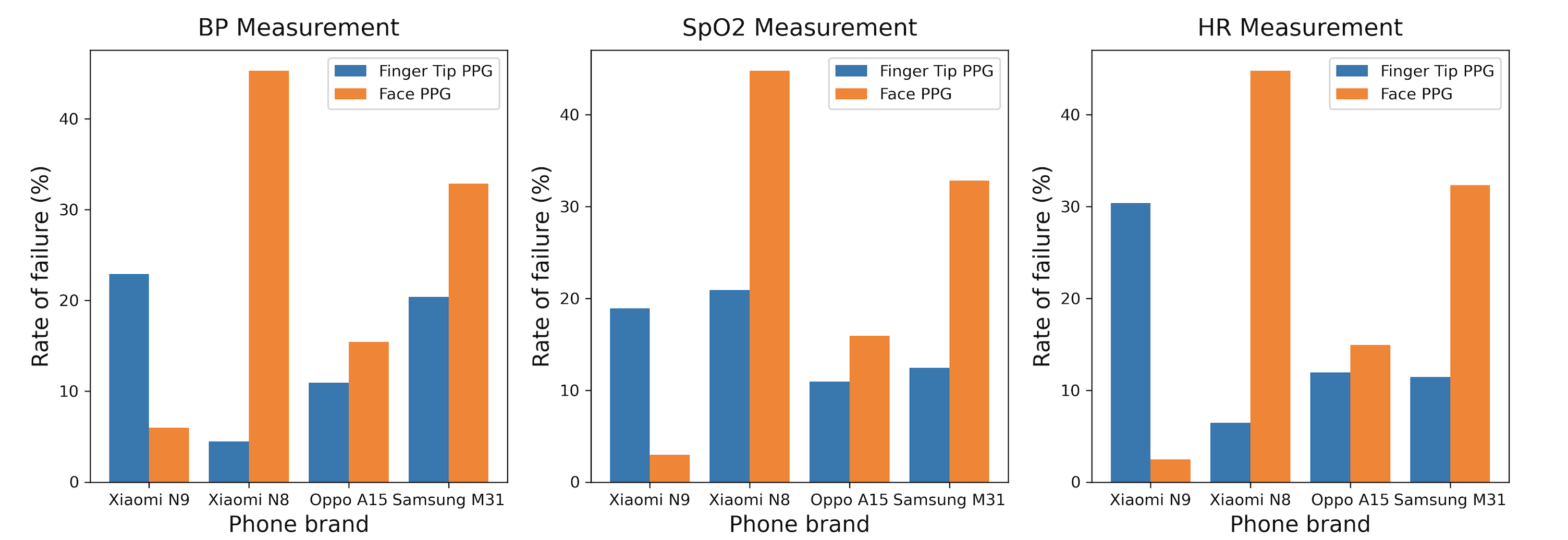}
    \caption{Failure rates in phones while measuring vitals using Finger tip PPG and Face PPG methods}
    \label{fig:Phone_Failure}
\end{figure*}

\begin{figure}
    \centering
    \includegraphics[width=3.2in]{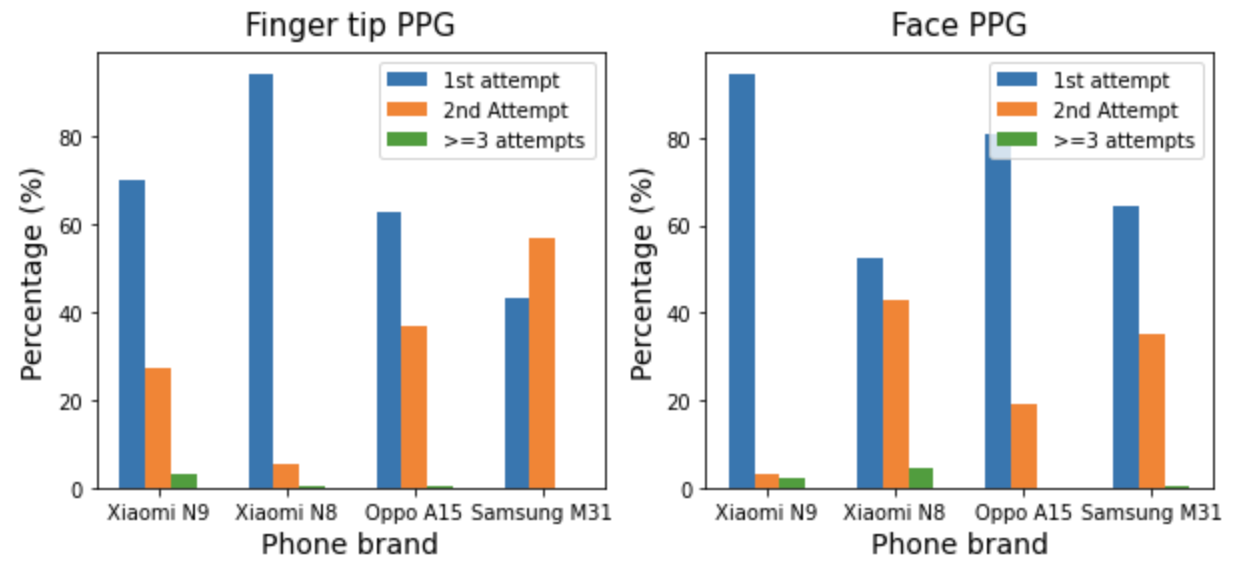}
    \caption{The number of attempts required to get a successful measurement of at least one vital}
    \label{fig:Attempt}
\end{figure}

%

\begin{table}[!t] \footnotesize
\centering
\renewcommand{\arraystretch}{1.3}
\caption{Error in camera based vitals measurement on mobile phone when readings of cardiac monitor are used as gold standard}
\begin{tabular}[!t]{c | c | c c c c } \hline
\cline{1-6}
  \multirow{2}{*} {Brand} & \multirow{2}{*} {Method} & { BP (S)} & { BP (D)} & SpO2 &  HR\\
     & & (mm/hg) & (mm/hg) & \bf (\%) & (/min)\\ \hline\hline
\multirow{4}{*}{ Xi N9} & \multirow{2}{*}{FT} & $\bf2.432$ & $\bf-0.680$ & $-2.277$ & $\bf-3.30$\\
 & & $\bf\pm 13.66$ & $\bf\pm10.25$ & $\pm3.28$& $\bf\pm 8.69$\\\cline{2-6}
 & \multirow{2}{*}{FC} & $-2.929$ & $-2.596$ & $\bf-1.368$ & $5.758$\\
 &  & $\pm 20.70$ & $\pm 13.60$ & $\bf\pm 2.32$& $\pm 14.70$\\\hline 		
\multirow{4}{*}{ Xi N8} & \multirow{2}{*}{FT} & $\bf1.783$ & $\bf-1.082$ & $\bf1.233$& $-5.919$\\ 	
 & & $\bf\pm 13.72$ & $\bf\pm 9.89$& $\bf\pm 3.25$& $\pm 10.13$\\\cline{2-6}
 & \multirow{2}{*}{FC} & $-2.929$& $-2.403$& $-1.526$& $\bf-2.103$\\
 &  & $\pm 18.76$ & $\pm 13.835$ & $\pm 2.406$& $\bf\pm 13.96$\\\hline
\multirow{4}{*}{ Oppo} & \multirow{2}{*}{FT}  & $\bf1.350$ & $-1.814$ & $\bf-1.5$ & $-7.70$\\
 & & $\bf\pm 13.62$ & $\pm 10.00$ & $\bf\pm 3.20$ & $\pm12.25$\\\cline{2-6}
 & \multirow{2}{*}{FC} & $-2.070$ & $\bf-0.824$ & $-1.543$& $\bf0.362$\\
&  & $\pm 17.251$ & $\bf\pm 12.937$ & $\pm 2.464$& $\bf\pm 14.91$\\\hline
\multirow{4}{*}{ Sm 31} & \multirow{2}{*}{FT}  & $\bf1.412$ & $\bf-1.412$ & $\bf0.866$ & $-4.797$\\
 & & $\bf\pm 13.90$ & $\bf\pm 10.24$ & $\bf\pm 5.00$ & $\pm 9.94$\\\cline{2-6}
 & \multirow{2}{*}{FC} & $-1.964$& $-3.157$ & $-1.105$&$\bf1.982$\\
&  & $\pm 18.10$ & $\pm 12.80$ & $\pm 3.621$& $\bf\pm 12.59$\\\hline
\end{tabular}
\begin{tablenotes}
\item Xi N9: Xiaomi Note 9 Pro; Xi N8:Xiaomi Note 8 Pro; Oppo: Oppo A15; Sm 31: Samsung M31; FT: Finger Tip PPG; FC: Face PPG; BP(S): Systolic Blood Pressure; BP(D): Diastolic Blood Pressure; SpO2: Blood Oxygen Saturation; HR: Heart Rate
\end{tablenotes}
\label{tab:phone_error}
\end{table}

\begin{table}[t] \footnotesize
\centering
\renewcommand{\arraystretch}{1.3}
\caption{Results of ANOVA on measurements of vitals across phones}
\begin{tabular}[!t]{c | c c c c } \hline
\cline{1-5}
  Method & { BP (S)} & { BP (D)} & SpO2 &  HR\\\hline\hline
FT (p-value) & $0.00002$ & $0.00002$ & $0.0$ & $0.067$\\
FC (p-value) & $0.967$ & $0.397$ & $0.520$& $0.04$\\\hline
\end{tabular}
\begin{tablenotes}
\item BP(S): Systolic Blood Pressure; BP(D): Diastolic Blood Pressure; SpO2: Blood Oxygen Saturation; HR: Heart Rate
\end{tablenotes}
\label{tab:anova_phone}
\end{table}

\begin{table}[t] \footnotesize
\centering
\renewcommand{\arraystretch}{1.3}
\caption{Error in vitals measurement on wearables when readings of cardiac monitor are used as gold standard}
\begin{tabular}[!t]{c | c c c c } \hline
\cline{1-5}
  \multirow{2}{*} {Brand} & { BP (S)} & { BP (D)} & SpO2 &  HR\\
      & (mm/hg) & (mm/hg) & \bf (\%) & (/min)\\ \hline\hline
\multirow{2}{*}{ GOQii} & $\bf2.015$ & $\bf-0.492$ & $-0.940$ & $\bf-4.422$\\
  & $\bf\pm 14.73$ & $\bf\pm11.58$ & $\pm3.725$& $\bf\pm10.53$\\\hline 		
\multirow{2}{*}{ Apple watch} & \multirow{2}{*}{ NA} & \multirow{2}{*}{ NA} & $\bf-0.744$& $-4.536$\\ 	
  &  & & $\bf\pm3.884$& $\pm 9.744$\\\hline
\end{tabular}
\label{tab:wearable_cm}
\end{table}

\begin{table}[t] \footnotesize
\centering
\renewcommand{\arraystretch}{1.3}
\caption{Results of ANOVA on measurements of vitals across wearables}
\begin{tabular}[!t]{c | c c } \hline
\cline{1-3}
  Method & SpO2 &  HR\\\hline\hline
p-value & $2.14$ & $0.928$\\\hline
\end{tabular}
\begin{tablenotes}
\item SpO2: Blood Oxygen Saturation; HR: Heart Rate
\end{tablenotes}
\label{tab:wearable_pval}
\end{table}

\begin{table}[!t] \footnotesize
\centering
\renewcommand{\arraystretch}{1.3}
\caption{p-value of differences in measurements between a device and the cardiac monitor using t-test}
\begin{tabular}[!t]{c | c | c c c c } \hline
\cline{1-6}
  \multirow{2}{*} {Brand} & \multirow{2}{*} {Method} & { BP (S)} & { BP (D)} & SpO2 &  HR\\
     & & (mm/hg) & (mm/hg) & \bf (\%) & (/min)\\ \hline\hline
{E} & - & $0.457$ & $0.457$ & $0.0009$ & $0.0009$\\\hline
{DM} & - & $0.537$ & $0.537$ & $0.0009$& $0.0009$\\\hline
\multirow{2}{*}{ Xi N9} & {FT} & $0.017$ & $0.780$ & $0.0$ & $0.011$\\\cline{2-6}
 & {FC} & $0.009$ & $0.121$ & $0.0$ & $0.00038$\\\hline 		
\multirow{2}{*}{ Xi N8} & {FT} & $0.269$ & $0.622$ & $0.0$ & $0.0001$\\\cline{2-6}
 & {FC} & $0.237$ & $0.279$ & $0.0$ & $0.784$\\\hline 
\multirow{2}{*}{ Oppo} & {FT} & $0.940$ & $0.005$ & $0.00001$ & $0.0$\\\cline{2-6}
 & {FC} & $0.015$ & $0.080$ & $0.026$ & $0.001$\\\hline 
\multirow{2}{*}{ SM31} & {FT} & $0.646$ & $0.098$ & $0.025$ & $0.002$\\\cline{2-6}
 & {FC} & $0.085$ & $0.008$ & $0.022$ & $0.751$\\\hline
{GO} & - & $0.033$ & $0.578$ & $0.0002$ & $0.0007$\\\hline
{AW} & - & - & - & $0.007$& $0.0002$\\\hline 
\end{tabular}
\begin{tablenotes}
\item Xi N9: Xiaomi Note 9 Pro; Xi N8:Xiaomi Note 8 Pro; Oppo: Oppo A15; Sm 31: Samsung M31; E: Expert; DM: Digital BP Monitor; GO: GOQii Smart watch; AW: Apple Watch; FT: Finger Tip PPG; FC: Face PPG; BP(S): Systolic Blood Pressure; BP(D): Diastolic Blood Pressure; SpO2: Blood Oxygen Saturation; HR: Heart Rate
\end{tablenotes}
\label{tab:all_comp}
\end{table}

Figure \ref{fig:Phone_Failure} shows the rate of failure in measuring the vitals on each of the phones considered for this experiment. It can be seen that Face PPG in general has a higher rate of failure compared to Finger tip PPG. While Finger tip PPG based methods had average failure rates of $14.676\pm7.383\%$, $15.796\pm4.197\%$ and $15.049\pm9.088\%$ for BP, SpO2 and HR, respectively, Face PPG based methods had average failure rates of $24.875\pm15.216\%$, $24.129\pm15.941\%$ and $23.631\pm16.169\%$. The plots in Fig. \ref{fig:Attempt} shows the number of attempts it took for those subjects where it was possible to successfully obtain a measurement. The percentage of users who could get successful readings in the 1st, 2nd and 3rd attempts for Finger tip PPG, respectively stood at $67.492\pm18.201\%$, $31.451\pm18.504\%$ and $1.056\pm1.241\%$, while for Face PPG it was $73.072\pm15.959\%$, $25.199\pm15.324\%$ and $1.728\pm1.658\%$ for 1st, 2nd and 3rd attempts, respectively.

Table \ref{tab:phone_error} shows that the error between the measurements obtained from Finger tip PPG based methods and cardiac monitor are in general lower compared to those obtained between Face PPG and cardiac monitor. However, the results of Finger tip PPG based measurements statistically varied across the phones for the same user while it remained quite similar for Face PPG across the phones as indicated by the p-values of Table \ref{tab:anova_phone}.


\subsection{Self monitoring of vitals using wearable}

There were no failures in measuring BP on GOQii and HR on both GOQii and Apple Watch. There was however a $2\%$ failure rate while measuring SpO2 on Apple Watch. The results of Table \ref{tab:wearable_cm} show that while GOQii smart watch has a lower error rate for HR, Apple Watch has a lower error rate for SpO2 when the results of cardiac monitor are used as the gold standard for comparison. The measurements between the two wearables were not statistically significant as is evident from Table \ref{tab:wearable_pval}. 

\subsection{Overall Comparison}
The results of Table \ref{tab:all_comp} shows that the inter-observer variability was statistically insignificant. While Finger tip PPG based measurements were in agreement with the gold standard for BP, the rest of the methods for all vitals showed statistically significant differences between the device measurements and gold standard.

\section{Discussion, Conclusion and Future work}
In this study we have shown that statistically significant variations exist in self monitoring of vitals using medical devices. One potential solution to address this is to sensitise users to the correct procedure to be followed while performing self-monitoring. An alternative to this would be to ask the subjects to use mobile based or wearable based solutions where the degrees of freedom for variabilities are fewer and easily manageable.  

The variabilities in self monitoring and monitoring with Mobiles and Wearables are similar. The ground truth measurements for algorithms on mobiles and wearables are obtained from experts who either use digital monitors or mercury based monitors (for BP) to measure vitals. Since an inherent variability existis in the technique used by experts as noted in \cite{villegas1995evaluation,schulze2002effect,ray2012blood}, this variability creeps into the training data for mobiles and wearables. What is even more interesting is the fact that the user variability in self monitoring and that by experts is more or less similar and thus the variabilities in self monitoring and that in mobiles and wearables are similar. A potential solution to eliminate this variability in training data will be to use the positional variability charts of Fig. \ref{fig:BP_Anatomy} and Fig. \ref{fig:SpO2_Anatomy} as reference while acquiring ground truth data for vitals. From the plots of Fig. \ref{fig:BP_Variations}, it can be seen that the position of the sensor in the BP cuff between [$315^{\text{o}}$,$22.5^{\text{o}}$] and [0cm,1cm] would result in less variability in BP measurements and thus result in consistent Ground Truth BP readings. Whereas, measurements taken with the hand in resting position with the index finger horizontally placed on the table is the ideal position to obtain Ground Truth measurements for SpO2 and HR.

Variabilities in hardware used for imaging within mobile phones result in statistically significant inconsistencies across mobile phones of different brand. Not only are the results inconsistent, it also takes multiple attempts on certain phones with certain technologies to get the measurements right. This will lead to bad user experience and loss of trust in the solution for vitals monitoring thereby resulting in a loss of adoptability of mobile camera based solutions for self-monitoring of vitals in the wild. One potential solution to reduce this variability will be to use camera calibration techniques which will normalize the color and white balance of the image stream or video that is being acquired for PPG signal construction.

Face PPG in general had a higher success rate compared to Finger tip PPG and that the varaibilty across phones for Face PPG was statistically insignificant compared to that of the Finger tip PPG based methods. This is a direct result of the experimental setup, where the phones were placed on a mobile holder for Face PPG and the user was asked to hold the phone for Finger tip PPG. This indicates the influence of the following two environment factors in the consistency of results: (a) lighting condition (the study was in a well lit room); and (b) motion artifacts. The impact of environmental factors as contributors for variability in self monitoring will be considered as an extension to this study.

\paragraph{Conclusion and Future Work}
In this paper we demonstrate the various variabilities that exist while performing self-monitoring of vitals using smart phones, wearables and medical devices and establish the statistical significance of the results of each when compared to the gold standard measurement obtained from a cardiac monitor. The study of environmental factors for variability in self monitoring, camera calibration for minimising hardware variability and PPG signal quality improvements will be extensions to our current work on Self monitoring in the wild using camera based solutions.
{\small
\bibliographystyle{ieee_fullname}

\bibliography{egbib}
}

\end{document}